\documentclass[10pt,conference]{IEEEtran} %
\IEEEoverridecommandlockouts
\usepackage{cite}
\usepackage{amsmath,amssymb,amsfonts}
\usepackage{algorithm}
\usepackage{ifthen}
\usepackage{multirow}
\usepackage{array}
\usepackage[noend]{algpseudocode}
\usepackage{graphicx}
\usepackage{textcomp}
\usepackage{xcolor}
\usepackage{xurl}
\usepackage{hyperref}
\usepackage{tikz}
\usepackage{svg}
\usetikzlibrary{shapes.geometric, arrows}
\def\BibTeX{{\rm B\kern-.05em{\sc i\kern-.025em b}\kern-.08em
    T\kern-.1667em\lower.7ex\hbox{E}\kern-.125emX}}
\begin{document}

\title{Towards Responsible AI in Education:\\ Hybrid Recommendation System for K-12 Students Case Study}

\author{
\IEEEauthorblockN{*Nazarii Drushchak}
\IEEEauthorblockA{
\textit{SoftServe Inc. and}\\
\textit{Ukrainian Catholic University}\\
Lviv, Ukraine \\
ndrus@softserveinc.com}
\and
\IEEEauthorblockN{*Vladyslava Tyshchenko}
\IEEEauthorblockA{
\textit{SoftServe Inc.}\\
Warsaw, Poland \\
vtysch@softserveinc.com}
\and
\IEEEauthorblockN{Nataliya Polyakovska}
\IEEEauthorblockA{
\textit{SoftServe Inc.}\\
Austin, Texas \\
npoly@softserveinc.com}
\thanks{*Nazarii Drushchak and Vladyslava Tyshchenko contributed equally to this work.}
}

\maketitle

\begin{abstract}
The growth of Educational Technology (EdTech) has enabled highly personalized learning experiences through Artificial Intelligence (AI)-based recommendation systems tailored to each student’s needs. However, these systems can unintentionally introduce biases, potentially limiting fair access to learning resources. This study presents a recommendation system for K-12 students, combining graph-based modeling and matrix factorization to provide personalized suggestions for extracurricular activities, learning resources, and volunteering opportunities. To address fairness concerns, the system includes a framework to detect and reduce biases by analyzing feedback across protected student groups.  This work highlights the need for continuous monitoring in educational recommendation systems to support equitable, transparent, and effective learning opportunities for all students.

\end{abstract}


\section{Introduction}

The rapid advancement of Educational Technology (EdTech) has significantly reshaped traditional learning environments, enabling the delivery of personalized educational experiences tailored to individual students’ needs. According to the U.S. Department of Education Office of Educational Technology, leveraging AI-based modern educational technologies has been pivotal in providing personalized pathways for learning, supporting adaptive and individualized instruction, and enhancing student engagement through innovative digital solutions\footnote{\url{https://tech.ed.gov/ai-future-of-teaching-and-learning/}}. This trend toward personalization in education underscores the importance of leveraging advanced recommendation systems to support student exploration and growth.

Recommendation systems in education have become critical in suggesting extracurricular activities, academic programs, and digital resources that align with students’ interests, aptitudes, and feedback. By leveraging these systems, educators can offer a more personalized learning experience, tailoring recommendations to each student’s unique profile and preferences. Such systems can positively impact student outcomes by fostering curiosity, encouraging skill development, and guiding students toward potential academic and career trajectories. However, despite their benefits, recommendation systems may also unintentionally introduce biases, resulting in disparate impacts on different student groups. For instance, prior research has highlighted potential biases in AI-based systems that can marginalize specific demographics, thereby limiting their access to valuable learning resources \cite{biasinedu}. 

In response to these concerns, this study introduces a graph-based recommendation system designed to provide personalized suggestions to K-12 students in public school districts. Independent software vendor SoftServe Inc. developed this solution for their client Mesquite Independent School District (ISD) as part of the creation of the personalized learning platform AYO\textsuperscript
{
\circledR
}. The primary goal of AYO\textsuperscript
{
\circledR
} is to harness data effectively to enhance student engagement and deliver tailored learning experiences that support each student's unique educational journey. The proposed system employs a hybrid approach, combining graph-based methods with matrix factorization to tailor recommendations based on students’ expressed and inferred interests. To promote responsible AI practices, we integrate a fairness analysis framework that systematically evaluates recommendations to identify and mitigate biases.

This work is structured as follows:

Section \ref{chap:relatedwork}
 provides an overview of the related work and background literature.
Section \ref{chap:researchgaps} identifies existing gaps in the integration of fairness-aware frameworks within graph-based recommendation systems in the educational domain.

Section \ref{chap:methodology}
outlines the design and implementation of the proposed hybrid graph-based recommendation system. It details the graph structure, integration of matrix factorization, and the development of the fairness analysis framework.

Section \ref{chap:casestudy}
presents a case study in a K-12 educational setting. It describes the dataset, experimental design, and implementation of the proposed system. The results are analyzed in terms of recommendation accuracy and fairness, highlighting the system’s effectiveness and any identified disparities.

Sections \ref{chap:fairnessmonitoring} and \ref{chap:biasmitigation} provide the approach to ensuring fairness within the recommendation system across demographic groups, identified by protected attributes such as gender, race, and socioeconomic status. The sections are focused on fairness monitoring and bias mitigation correspondingly. 

The study’s conclusions are presented in section \ref{chap:conclusion}, highlighting key findings and their implications.

Section \ref{chap:worklimitations} discusses the study’s limitations and suggests directions for future research.

Ethical aspects of data privacy, informed consent, and legal compliance are addressed in section \ref{chap:ethicalconsiderations}.

\section{Background and Related Work}
\label{chap:relatedwork}
\subsection{Personalization in Educational Technology}

Personalized learning systems in EdTech leverage AI-based recommender systems to tailor content based on student preferences. Li and Chen \cite{recommendationsystemsforadaptivelearning} demonstrated the effectiveness of these systems in improving engagement and outcomes. Additionally, a systematic literature review on educational recommendation systems \cite{edurecsystems} was conducted, highlighting trends in recommendation production, evaluation methods, and research gaps. Their findings indicate that hybrid approaches dominate, but evaluations often focus solely on accuracy, neglecting the pedagogical impact. This underscores the need for multidimensional evaluation frameworks to better assess the effectiveness of these systems in supporting teaching and learning activities.

\subsection{Graph-Based Recommendation Systems}
Graph-based approaches are valued for their ability to model complex user-resource interactions. There was demonstrated the efficacy of Graph Convolutional Networks (GCNs) in capturing large-scale relationships \cite{gccn}.

In addition to traditional applications, graph-based recommendation methods have shown considerable promise in specialized domains like the academic community \cite{8691158}. Furthermore, the researchers \cite{wang2021graphlearningbasedrecommender} provide a comprehensive review of graph learning-based recommendation systems, discussing various methodologies and highlighting their adaptability across diverse use cases. 

\subsection{Fairness in AI-Based Recommendation Systems} 
While personalized recommendation systems have shown the potential to improve learning experiences, concerns about fairness and algorithmic bias remain prominent. Studies \cite{biasinedu} have drawn attention to the risks of reinforcing existing inequalities in educational settings.
Binns \cite{individual_fairness} and Burke \cite{burke2017multisidedfairnessrecommendation} proposed fairness-aware frameworks to address these issues.
Recommendation systems have a different logic than traditional machine learning tasks, as they rely on user-item interactions and dynamic feedback loops, making it not optimal to assess fairness using standard metrics. Authors of \textit{``Fairness in Recommendation Systems: Research Landscape and Future Directions''} work\cite{Deldjoo2024} have reviewed current studies on fairness in recommendation systems, highlighting various issues and gaps in existing methods.
In addition to this, algorithmic fairness, the main component of responsible AI, has been analyzed comprehensively, with fairness defined in various ways based on philosophical considerations and contextual use \cite{khan2022substantive}. Researchers have developed numerous fairness metrics to address different aspects of fairness \cite{bird2020fairlearn,aif360-oct-2018,saleiro2018aequitas,DBLP:journals/bigdata/Chouldechova17,friedler2019comparative,10.1145/3457607,verma2018fairness}.

\subsection{Hybrid Approaches in Recommendation Systems} 
Hybrid recommendation systems have been widely adopted to combine the strengths of different recommendation techniques.
Koren et al. \cite{5197422} introduced matrix factorization techniques that uncover latent patterns in user-item interactions, which have become foundational in collaborative filtering approaches.

For instance, there was proposed a scalable and accurate hybrid recommendation system that combines collaborative filtering with content-based filtering \cite{5432716}. 

Similarly, in the educational space, there was developed a personalized recommendation system for college libraries that combines collaborative filtering and content-based techniques to help users navigate vast collections of books \cite{TIAN2019490}.

\section{Research Gaps and Problem Formulation}
\label{chap:researchgaps}

While existing studies \cite{recommendationsystemsforadaptivelearning,edurecsystems,Deldjoo2024} provide valuable insights into personalized learning and fairness-aware recommendation systems, there remains a need for approaches that can effectively combine graph-based methods with fairness analysis in the educational domain. Current research often treats recommendation generation and fairness analysis as separate processes, resulting in challenges when trying to achieve real-time fairness monitoring and bias mitigation. Moreover, many traditional approaches \cite{wang2021graphlearningbasedrecommender,5197422,5432716,TIAN2019490} focus primarily on optimizing quality, with limited emphasis on transparency and interpretability—both of which are critical in educational applications to ensure trustworthiness and accountability. 

To address these issues, our work seeks to combine graph-based modeling and matrix factorization with a fairness analysis framework. This integration aims to enhance both the accuracy and equity of recommendations, paving the way for more personalized and responsible AI-driven learning experiences for K-12 students.

\section{Methodology}
\label{chap:methodology}
Our study focuses on the fairness analysis of a graph-based recommendation system \cite{9216015}. The recommendation system provides personalized suggestions for students based on their interests, aptitudes, and explicit feedback, while the fairness analysis ensures that these recommendations do not introduce biases or unfair treatment toward any group.

\subsection{Graph Recommendation System}
Our graph-based approach for the recommendation system supports responsible AI practices by being transparent, integrating diverse data, and effectively capturing relationships between interests, aptitudes, and recommended targets while maintaining interpretability.

The graph structure in our system consists of two parts: 
\begin{itemize} 
    \item \textbf{Static Part} defines relationships between interests, aptitudes, and various resources. The edges represent the cosine similarity between the descriptions of interests, aptitudes, and resources. To generate embeddings, we use the \textit{Sentence Transformer}\footnote{\url{https://www.sbert.net/}} library and \textit{paraphrase-multilingual-mpnet-base-v2}\footnote{\url{https://huggingface.co/sentence-transformers/paraphrase-multilingual-mpnet-base-v2}} model, which allows us to capture semantic similarities across different languages and contexts. 
    \item \textbf{Dynamic Part} contains edges between individual students and their identified interests and aptitudes. The weight of those edges is based on students' explicit feedback and algorithms defined in other parts of the educational platform.
\end{itemize}

The recommendation process is hybrid and consists of both recommendations based on graph neighborhood and matrix factorization. Students can provide explicit feedback on each recommendation they receive, such as indicating the relevance or usefulness of the suggested resource. This feedback is then incorporated into the system to refine future recommendations and is also used for the fairness audit process.

\subsection{Fairness Analysis}
Fairness analysis in recommender systems is crucial for ensuring that the suggestions provided do not inadvertently discriminate against certain groups of users. In this context, protected groups refer to defined categories of individuals who may face discrimination (for example, gender, race, and family status), while protected attributes are specific characteristics associated with those groups, as outlined by anti-discrimination laws\footnote{\url{https://www.justice.gov/crt/federal-protections-against-national-origin-discrimination-1}}.

For our recommendation system, we propose a concept of fairness analysis based on user reactions. Our approach involves evaluating fairness by analyzing positive and negative reactions across different resources and protected groups. The process includes:

\begin{enumerate} 
    \item Data collection: Gathering feedback from users, and categorizing it into positive and negative reactions.
    \item Segmentation by protected groups: Analysis of the feedback for each type of resource, separately considering each protected attribute across different protected groups (e.g., gender, race, family status).
    \item Comparison of feedback: Comparison of the percentage of positive and negative reactions across diverse groups to identify disparities.
    \item Bias alerting and auditing: Utilization of the feedback analysis as an alerting mechanism. This analysis enables us to pinpoint specific resources or recommendations that may pose problems and clarify the reasons for these issues within particular user subsets, which can then inform targeted mitigation strategies.
\end{enumerate}

It is important to note that this process serves primarily as a tool for alerting and auditing rather than a complete solution. It helps to detect potential issues, after which further actions are needed to address and mitigate these biases.

\section{Case Study}
\label{chap:casestudy}
\subsection{Solution Explanation}
Our hybrid recommendation system is developed as a part of an educational platform aimed at providing personalized learning experiences for public school districts' K-12 students. The recommendation system suggests a unique set of opportunities for each student to help them grow their potential. Outputs of the recommendation system may point students to the exploration of new interests, development of novel or building up of existing skills, meaning that each suggestion from the system may impact student's future. Hence, the system needs to be equally safe, trustworthy, and ethical to each of its young users. Our system is designed with the notion of Responsible AI principles \cite{trustworthy_ai}, making sure that fairness, reliability, and transparency are the core features of the system from the launch day.   

\subsubsection{Transparency}
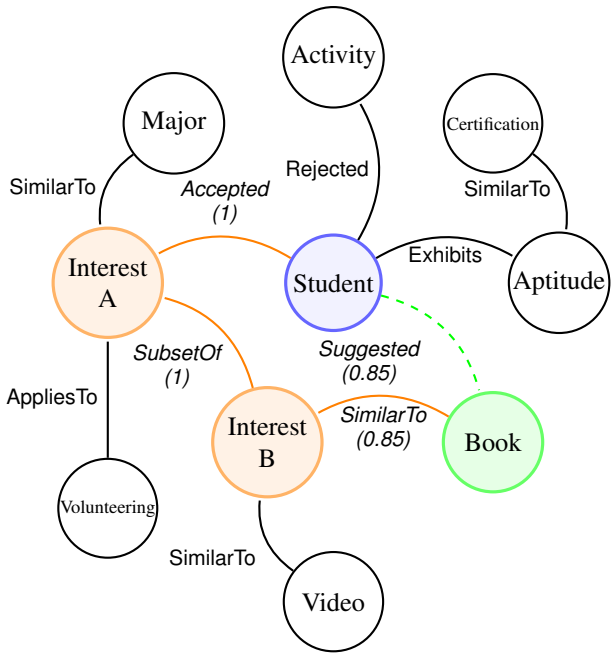
\begin{figure}
    \centering
    \begin{tikzpicture}[>=stealth',shorten >=1pt,auto,node distance=3cm,
                        thick,main node/.style={circle,minimum size=13mm, text width=13mm,inner sep=0pt,draw,align=center},
                        student node/.style={circle, draw=blue!60, fill=blue!5, very thick, minimum size=12.5mm, text width=12.5mm, inner sep=0pt,draw,align=center},
                        suggestion path node/.style={circle, draw=orange!60, fill=orange!10,very thick, text width=13mm,inner sep=0pt,draw,align=center},
                        suggested node/.style={circle, draw=green!60, fill=green!10,very thick, text width=13mm,inner sep=0pt,draw,align=center},
                        edge label/.style = {inner sep=1pt}
                        ]
    
      \node[student node] (1) {Student};
      \node[suggestion path node] (2) [left of=1] {Interest A};
      \node[main node] (3) [right of=1] {Aptitude};
      \node[main node] (4) [above of=1] {Activity};
      \node[suggested node] (7) [below right of=1] {Book};
      \node[main node] (6) [above left of=1] {Major};
      \node[main node] (8) [below left of=7] {Video};
      \node[main node] (9) [below of=2] {\scriptsize Volunteering};
      \node[main node] (5) [above right of=1] {\scriptsize Certification};
      \node[suggestion path node] (10) [below right of=2] {Interest B};
    
      \path[every node/.style={font=\sffamily\footnotesize}]
        (1) edge [bend right, orange] node[above, black] {\textit{\begin{tabular}{c}  Accepted \\ (1) \end{tabular}}} (2)
            edge [bend right] node[left] {Rejected} (4)
            edge [bend left] node[below] {Exhibits} (3)
            edge [bend left, green, dashed] node[below left, black] {\textit{\begin{tabular}{c}  Suggested \\ (0.85) \end{tabular}}} (7)
        (9) edge node[left] {AppliesTo} (2)
        (10) edge [bend right, orange] node[edge label, near start, left, black] {\textit{\begin{tabular}{c} \\ SubsetOf \\ (1) \end{tabular}}} (2)
        (8) edge [bend left] node {SimilarTo} (10)
        (6) edge [bend right] node[left] {SimilarTo} (2)
        (7) edge [bend right, orange] node[below, black] {\textit{\begin{tabular}{c}  SimilarTo \\ (0.85) \end{tabular}}} (10)
        (5) edge [bend left] node[left] {SimilarTo} (3);
    \end{tikzpicture}
    \caption{\textbf{Representation of Student Neighbourhood Subgraph}\newline
    Sample representation of induced subgraph of neighbors centered at Student node within a given radius. Student nodes (blue) represent each unique student and connect with other types of entities. Orange nodes represent a path for new suggestions (neighborhood of radius 3). A green node represents a new suggestion and new connection in the graph of type "Suggested". Each edge in the graph has type and weight (0;1] located under the edge label)}
    \label{fig:student_subgraph}
\end{figure}

    \paragraph{Content-filtering part}
      To prevent a cold-start problem \cite{cold_start_problem} of a new community, we employ a content-filtering approach based on undirected weighted graphs. We store graph nodes and edges in tabular format for cost efficiency purposes and use the \textit{NetworkX}\footnote{\url{https://networkx.org/}} library to initialize graphs for each student in a distributed data-parallel fashion using \textit{Apache Beam SDK}\footnote{\url{https://beam.apache.org/}}. When the student graph is initialized, it is reduced to the subgraph with a certain neighborhood radius. Figure \ref{fig:student_subgraph} provides a sample schematic representation of a neighborhood subgraph for one student, highlighting different types of entity nodes and edge connections.  Suggestions are then selected based on Dijkstra’s shortest path algorithm \cite{dijkstra} with the Student node as a source. Suggestions ranks are defined based on the multiplication of shortest path weights and number of shortest simple paths between the Student node and suggestion node. The key feature that allows solution transparency is logging the reasoning behind each suggestion, which includes a list of nodes in the shortest path and ranks as a suggestion confidence score.

\begin{figure}[h!]
            \includegraphics[width=\columnwidth]{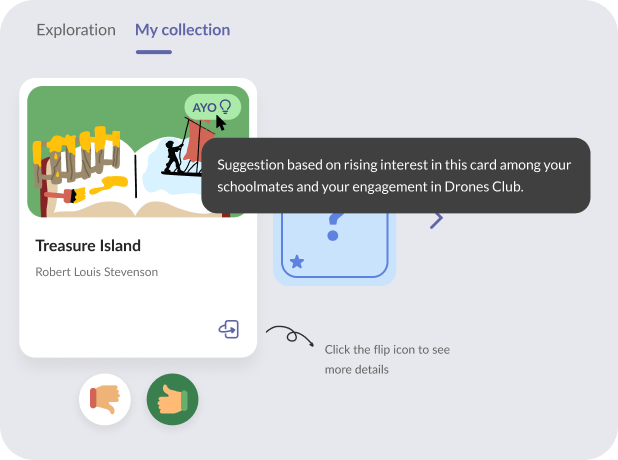}
            \caption{\textbf{Element of the Recommendation System User Interface in the Solution} \newline 
            Whenever a user reviews a new recommendation, the hover over the suggestion icon triggers an appearance of the tooltip with the reasoning behind the recommendation.}
            \label{fig:ui_reactions}
        \end{figure}

      \paragraph{Collaborative-filtering part}
      To effectively utilize explicit users' reactions for high-quality recommendations we also use a collaborative-filtering approach based on the Non-Negative Matrix Factorization algorithm \cite{6795860} from the \textit{Scikit Learn}\footnote{\url{https://scikit-learn.org/dev/modules/generated/sklearn.decomposition.NMF.html}} library. Positive reactions are used to build a users-targets matrix, while negative reactions suit for filtering out redundant suggestions. Weights of the reconstructed matrix are used as confidence scores and are logged into the system for each suggestion.
      \newline\newline
      Both content-filtering and collaborative-filtering recommendations are later re-ranked in case some targets were selected by both methods. All confidence scores and reasoning behind suggestions are accumulated in the system's database for debugging purposes. They are also later used for fairness audits. Moreover, a user interface of the system provides the reasoning behind each suggestion according to \textit{Human Interface Guidelines for Machine Learning Applications}\footnote{\url{https://developer.apple.com/design/human-interface-guidelines/machine-learningl}}. The element of the solution's user interface is displayed in Figure \ref{fig:ui_reactions}, highlighting the incorporation of the transparency principle into the user's experience.

\subsubsection{Reliability}
The reliability component of our system is guaranteed and measured within two realms: content reliability and recommendations quality.
\paragraph{Content Reliability}
The content of our recommendation system has multiple origins: internet-sourced resources (Courses), ML-generated resources (Activities), and district resources (Books, Videos, Extracurriculars, Certifications, and Volunteering activities).
\begin{itemize} 
    \item \textit{Internet-sourced} resources include free courses sourced from platforms like Coursera \footnote{\url{https://www.coursera.org/}}.   All sourced courses are manually verified by human moderators to ensure the age-level appropriateness of each recommended course.
    \item \textit{ML-generated} resources include short Activities generated by GPT-2 model \cite{gpt2}. All generated items were manually verified by human moderators to ensure the reliability of generated content.
    \item \textit{District} resources include materials used by the school district in their libraries. Due to their amount, it is not feasible to verify them manually. We use Google Cloud Natural Language Text Moderation API \footnote{\url{https://cloud.google.com/natural-language/docs/moderating-text}} and PaLM2 model \cite{anil2023palm2technicalreport} with prompt engineering to filter out potentially unsafe resources using their descriptions.
\end{itemize}

\paragraph{Recommendations Quality}
To continuously monitor the quality of the recommendations, we calculate multiple evaluation metrics used for information retrieval algorithms. A list of the main evaluation metrics is available in Table \ref{table:evalution_metrics}. Each of the metrics is aggregated by target category and by grade levels (elementary, middle, high school) to differentiate the evaluation process for different subgroups of users. All evaluation metrics are supported by real-time visualizations which are integrated into the district's Looker \footnote{\url{https://cloud.google.com/looker}} dashboard.

\begin{table}[h!]
    \begin{center}
    \caption{Evaluation Metrics}
    \label{table:evalution_metrics}
        \begin{tabular}{  |m{10em} | m{18em}|}
         \hline
         \textbf{Metric} & \textbf{Intuition} \\
         \hline
         Coverage & Ratio of targets that have been recommended to at least one user\\  
         \hline
         Precision & Ratio of recommendations that have received positive feedback from users \\  
         \hline
         Mean Average Precision & Average Precision at various cutoff levels to evaluate the ranking of recommendations \\ 
         \hline
         Personalization Rate & Average level of uniqueness of suggestions among users  \\  
         \hline
        \end{tabular}
    \end{center}
\end{table}

\subsubsection{Fairness}
To guarantee the fairness of our recommendation system, we developed a fairness audit procedure that measures the system's fairness across chosen protected variables. Specifically, we measure statistical fairness given ground truth through equalized odds, described by Wang et al. \cite{brief_review_algorithmic_fairness}.
We formally define the fairness of our system as a \textit{function of the difference between recommendations precision} of the two or more protected groups. We focus on precision as both True Positive (TP) and False Positive (FP) recommendations are considered during its calculation. Explicit positive and negative user feedback is used as a ground truth. \newline\newline Table  \ref{table:protected_list} displays chosen protected variables and groups along with additional explanations. It is important to note that the choice of protected variables and groups was defined by the district's student management system and data attribute availability. 
\begin{table}[h!]
    \begin{center}
    \caption{Protected variables and groups}
    \label{table:protected_list}
        \begin{tabular}{  |m{9em} | m{8em}| m{9em} | }
         \hline
         \textbf{Protected Variables} & \textbf{Protected Groups} & \textbf{Explanation} \\ 
         \hline
         Gender & F (female), M (male) & - \\  
         \hline
         Race Code & 1-5  & - \\  
         \hline
         Has Parents & mother only, father only, mother and father, other & - \\
         \hline
         Is Homeless & binary & true if student is considered homeless\\
         \hline
         Is Migrant & binary & true if student migrated from another state\\
         \hline
         Is Immigrant & binary & true if student migrated from another country\\
         \hline
         Is Foster & binary & true if a student is in foster care \\
         \hline
         Is Gifted & binary & true if a student is a member of Gifted/Talented program \\
         \hline
        \end{tabular}
    \end{center}
\end{table}

Our fairness audit procedure is conducted for each protected variable in Table \ref{table:protected_list} separately.
Algorithm \ref{alg:fairness_audit} outlines the main steps taken in this procedure. For all pairs of protected groups $g_j$, $g_{j+1}$ we calculate the variations in fairness outcomes as differences in recommendations precision $\left| P(g_j, t_i) - P(g_{j+1}, t_i) \right|$ within the chosen target  $t_i$.
We also define $\Delta{P}$ as the tolerance level  \cite{fairness_tolerance} at which fairness variations are considered acceptable.
If the value exceeds the tolerance level $\Delta{P}$, pairs of $(g_j, t_i)$ and $(g_{j+1}, t_i)$ are flagged for further analysis of user reactions and reasoning behind suggestions.  Our setting for tolerance level $\Delta{P}$ is set to 0.1, relaxing the standard 0.05 probability of false rejection \cite{Adel_Valera_Ghahramani_Weller_2019} used for classification systems. To ensure that the fairness audit is conducted on a group level rather than an individual, we set a threshold $N_{sample}$ which indicates the sufficiency of reactions sample for each protected group. $N_{sample}$ is set to 10 "active" users per group, meaning that to be considered for fairness audit, the protected group should have at least this number of unique users with reactions.

\begin{algorithm}
    \caption{Fairness Audit Procedure}\label{alg:fairness_audit}
    \begin{algorithmic}[1]
            \Procedure{Fairness Audit}{}
            \State $\textit{V} \gets \text{protected variable (e.g., gender)}$
            \State $\textit{G} \gets \text{set of protected groups within \textit{V}}$
            \State $\textit{T} \gets \text{set of targets (e.g., book, video, certification)}$
            \State $N_{sample} \gets \text{min required number of active users}$
            \State $\Delta{P} \gets \text{tolerance level of variations in fairness}$
            \For {$t_i$ \text{in} \textit{T}}
                \For {$g_j$ \text{in} \textit{G}}
                    \If{$len(\text{active users}_{g_j, t_i}) \ge N_{sample}$}
                      \State calculate $P_{g_j,t_i}$ \Comment{$P$ = Precision}
                    \Else { ignore calculation for $(g_j, t_i)$}
                    \EndIf
			\EndFor
		      \If{$\left| P(g_j, t_i) - P(g_{j+1}, t_i) \right| \leq \Delta{P} $,  $ \forall g $}
                  \State recommendations within $t_i$ are fair
                \Else {}
                  \State compare reactions of $(g_j, t_i)$ and $(g_{j+1}, t_i)$
                  \State compare the reasoning of $(g_j, t_i)$ and $(g_{j+1}, t_i)$
                \EndIf
            \EndFor
            \EndProcedure
    \end{algorithmic}
\end{algorithm}

We believe that the setting for the fairness audit of our system is multipurpose. Although its primary goal is to measure fairness, during the analysis of variations in recommendations precision we were also able to verify the transparency of the system (how deep root cause analysis it allows for) as well as its reliability (are any of the variations in fairness caused by unreliable content). 

\subsection{Fairness Audit Results}

To effectively analyze variations in fairness across protected groups and variables in Table \ref{table:protected_list} we built visualizations that complement Algorithm  \ref{alg:fairness_audit}. This section is focused on the fairness analysis of \textit{gender} protected groups specifically. We conducted an equivalent audit and analysis for all other protected variables of interest. %

\begin{figure}[p]
    \includegraphics[width=\columnwidth]{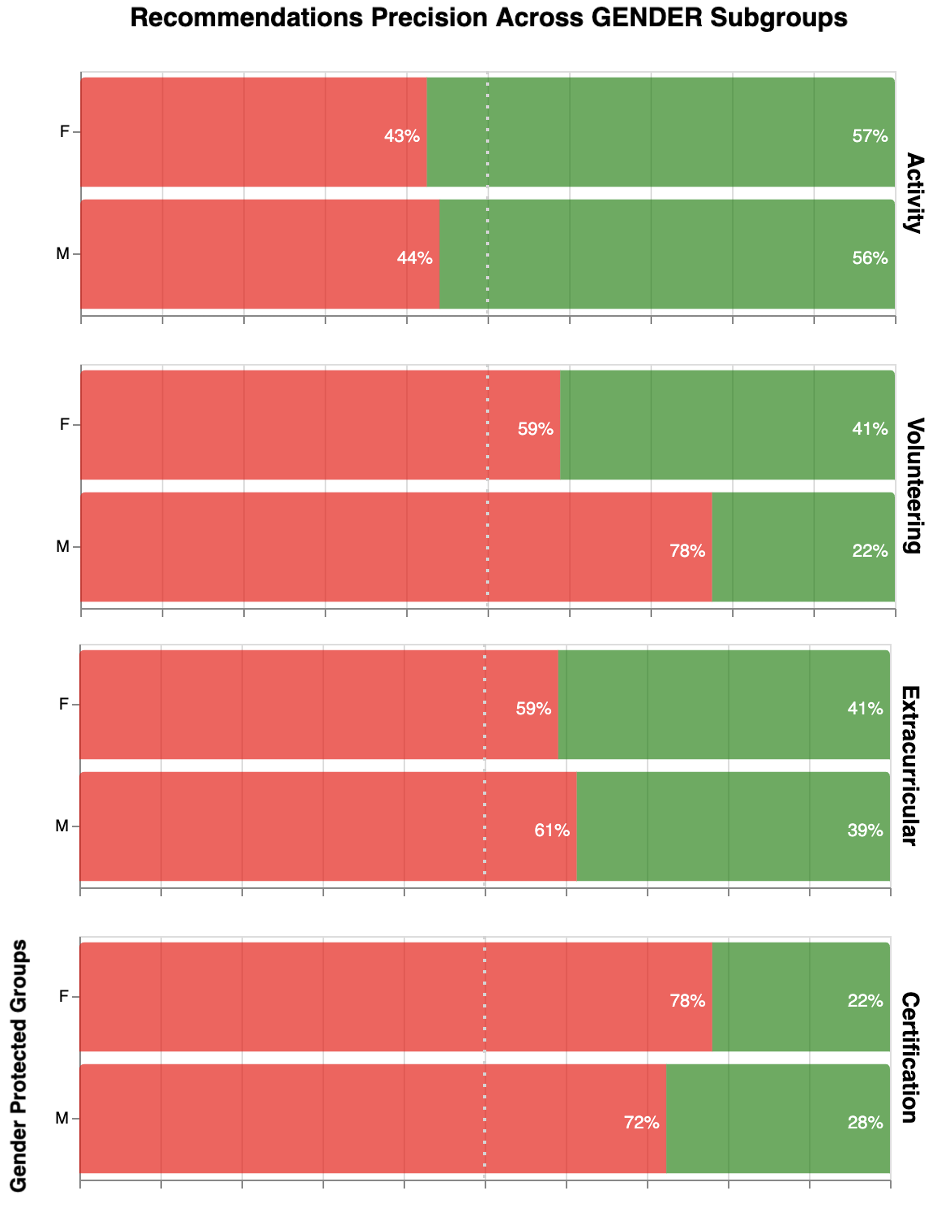}
    \includegraphics[width=\columnwidth]{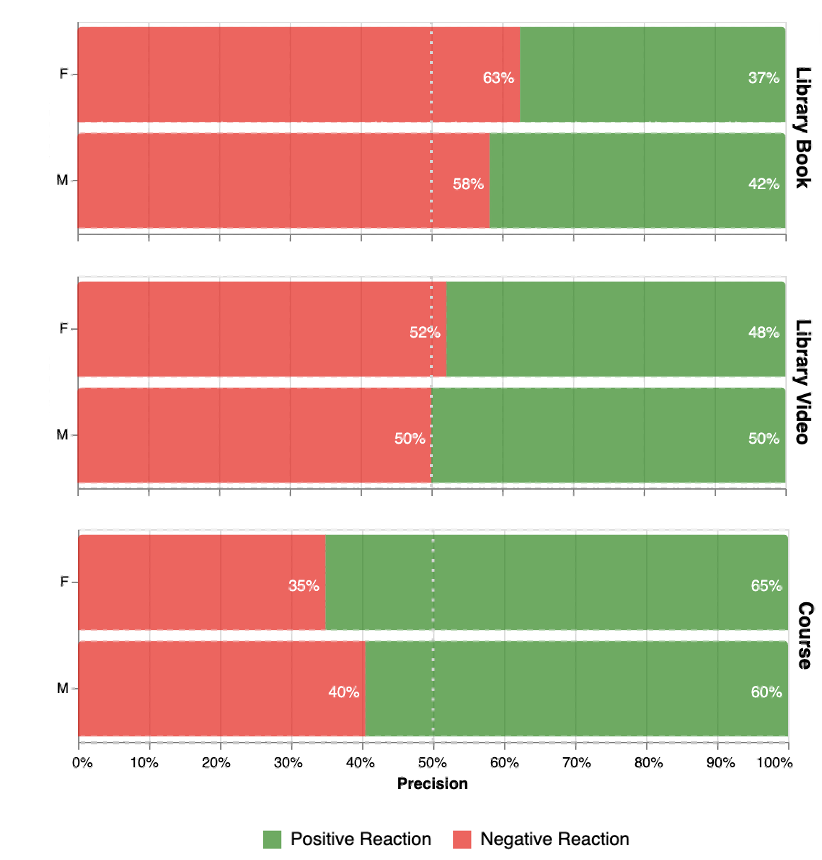}
    \caption{\textbf{Recommendations Precision In Gender Groups} The Protected group label (F/M) is displayed on the left side of the vertical axis. The target category is shown on the left side of the vertical axis. The horizontal axis of each chart provides recommendations for precision percentage.}
    \label{fig:gender_fairness_charts}
\end{figure}

An outlook on the precision of recommendations in gender subgroups is provided in Figure \ref{fig:gender_fairness_charts}. The figure shows that the only target category that exceeds the set tolerance level is Volunteering. It is also notable that the proportion of negative and positive reactions is not equal between targets, for example, both protected groups have higher satisfaction in Activities recommendations than in Certification. Although this aspect was analyzed by us during the audit, in this work we focus on variations in fairness rather than differences in precision between targets. \newline
We further explore the possible causes of high variation in precision for Volunteering between gender-protected groups by comparing:\newline
\begin{enumerate}
  \item \textbf{Ranks of top negative and positive reactions}. The only targets with negative reactions ranked higher by the male gender group are \textit{cleaning up cigarette butt litter to protect humans, animals, and the planet} - 5 reactions,  \textit{and keep mesquite beautiful community cleanup} - 4 reactions.
  \item \textbf{Unique negative and positive reactions}. The are no targets that have negative reactions only by male gender group.
  \item \textbf{The reasoning behind suggestions with negative reactions} is based on suggestions path from the graph or weights in a reconstructed matrix.  To target \textit{clean up cigarette butt litter to protect humans, animals, and the planet} graph suggestions paths are: \texttt{an aptitude for organizational}, \texttt{aptitude physical}, \texttt{and interest ecology environments}. For target \textit{keep Mesquite beautiful community cleanup}  - \texttt{aptitude organizational}, \texttt{aptitude physical}, \texttt{interest environmental issues} respectively. Suggestions paths are the same for both gender groups for these two targets.
\end{enumerate}
$\newline$ Based on the provided results, we can state that both recommendations and their reasoning for the Volunteering target are not biased towards any gender group. All suggestions paths are valid, signifying that graph nodes and edges involved in the recommendation of the mentioned targets are gender-neutral. Hence, the increased difference in recommendations precision should be explained by other factors, for example: 
\begin{enumerate}
  \item Gender bias in interest or aptitude identification algorithms.
  \item Insufficient sample size that should be higher and focus on group-target level, rather than just on the group.
  \item Other hidden factors, that are out of the scope of the recommendation algorithm (venue location, activities schedule co-occurring with other campus events, etc.).
\end{enumerate}
$\newline$ For all other targets, we received either similar results on variation in fairness or no variation at all. Variables and targets flagged for additional analysis are summarized in Table \ref{table:fairness_results}. Protected variables that are not displayed in Table \ref{table:fairness_results} proved not to exceed the tolerance level for all target categories. The visualizations for all target categories and protected groups are available in our GitHub repository. \footnote{\url{https://github.com/mesquiteisd/rai-hybrid-recommendation-k12.git}} 

\begin{table}[h!]
    \begin{center}
    \caption{Protected Variables and Targets Flagged for Analysis}
    \label{table:fairness_results}
        \begin{tabular}{  |m{8em}|m{8em}|m{6em}| }
         \hline
         \textbf{Protected Variable} & \textbf{Target} & \textbf{Variation in Fairness} \\ 
         \hline
         Gender & Volunteering & 0.19 \\  
         \hline
         \multirow{2}{8em}{Race Code} & Extracurriculars  & 0.12 \\  
          & Library Video & 0.19 \\
         \hline
         Is Immigrant & Library Video & 0.15\\
         \hline
         \multirow{2}{8em}{Has Parents} & Volunteering & 0.14\\
          & Certification & 0.18\\
          & Library Book & 0.16\\
         \hline
        \end{tabular}
    \end{center}
\end{table}

\section{Recommendations for Fairness Monitoring}
\label{chap:fairnessmonitoring}
To ensure fairness in recommendation systems similar to ours, we emphasize the importance of regular fairness audits together with the monitoring of the system's reliability. According to the ``Fairness in
Recommendation: Foundations, Methods and Application'' study \cite{li2023fairnessrecommendationfoundationsmethods}, there are many expressions of fairness in the taxonomy of fairness notions in recommendation systems. Developers should recognize which categories are most relevant for their decision-making systems and target them in the audits. Audit results should be continuously shared with stakeholders, ensuring transparency about any identified issues and the steps are taken to address them. Lastly, allowing users to provide explicit feedback including provisioning fairness concerns, will make it easier to recognize and correct any unfair treatment for cases that may not be covered by fairness audits.

\section{Recommendations for Bias Mitigation}
\label{chap:biasmitigation}
Although the results above proved that no action should be currently taken to mitigate bias in our recommendation system, we consider multiple fairness-aware mechanisms that could be implemented in similar systems. \newline
\textit{Pre-Processing Mechanisms:} 
\begin{itemize} 
    \item Data re-sampling. Biased targets (e.g., assume belonging to one of the protected groups) should be removed from the system or their amount should be equal for each protected group. Such targets can be identified via classification using Large Language Models.
\end{itemize}
\textit{In-Processing Mechanisms:}
\begin{itemize} 
    \item Introducing fairness constraints into the recommendation process. Protected group labels can be injected as separate graph nodes while weights of the edges between protected groups and targets should be regularized to balance the recommendations.
\end{itemize}
\textit{Post-Processing Mechanisms:}
\begin{itemize} 
    \item 
    Non-parametric re-ranking. Under specified fairness constraints, the optimal re-ranking outcomes can be found through heuristic search methods \cite{JIN2023101906}. Specifically, given original top-k recommendation results for each user from the system, heuristic methods maximize the total preference score concerning fairness by rearranging original recommendations. 
\end{itemize}

\section{Conclusion}
\label{chap:conclusion}
This paper presented a hybrid recommendation system as a case study based on a real-world solution developed by SoftServe Inc. for Mesquite Independent School District  \footnote{\url{https://www.softserveinc.com/en-us/education-reimagined}}. Our research highlights the critical need for fairness analysis in educational systems, particularly in recommendation systems that can influence student opportunities and outcomes. By incorporating a fairness audit framework into the design of our system, we aim to identify and mitigate potential biases, ensuring that all students have equitable access to educational resources. 
Future efforts will aim to enhance the robustness of our fairness evaluation methods and explore advanced mechanisms to ensure that the recommendations provided are not only personalized but also equitable for all students.

\section{Work Limitation}
\label{chap:worklimitations}

Our study has a few limitations, which we plan to address in future work:

\begin{itemize}
    \item Simple feedback mechanism: We rely only on explicit positive/negative feedback from users.
    \item Single-attribute fairness analysis: Our fairness evaluation only considers a single protected variable per analysis, not a combination of groups (e.g., race\&gender, gender\&migrant).
    \item Single fairness metric: The study focuses on recommendations precision as a main fairness metric. Our fairness audit does not cover the fairness of recommendations ranking.
    \item Manual bias correction: Fairness audits were conducted manually, with no automated alerting mechanism.
    \item In-house fairness audit: The analysis was conducted by the development team, risking bias.
    \item Limited data on protected groups: Protected groups covered in fairness audit may not fully represent the whole population of the users. The choice of protected variables and their groups is solely based on the availability of records in the students' management system.
\end{itemize}

\section{Ethical Consideration}
\label{chap:ethicalconsiderations}
In this study, we prioritize fairness, aiming to identify and address potential biases in hybrid recommendation systems. Our research intends to promote equitable recommendations by raising awareness of these biases. We acknowledge the responsibility to handle this sensitive topic with care and strive to contribute positively to the discourse on fairness and equity in recommendation systems. Data used in this study was obtained from a school district and has been fully anonymized. All students and their parents provide consent for the use of students' data in the system. Additionally, we have ensured that data can be promptly deleted upon request from any participant. Throughout the paper, we present only aggregated results, ensuring that no individual's data can be identified.
 
We used ChatGPT\footnote{\url{https://chat.openai.com/}} and Grammarly\footnote{\url{https://www.grammarly.com/}} to aid in paraphrasing while writing this work, ensuring that our language is clear and respectful.

\section*{Acknowledgments}
\label{chap:acknowledgments}
We would like to thank the development team of AYO\textsuperscript
{
\circledR
} for collaborating with us on this solution. We are also thankful to SoftServe Data Science Group for their support and to the administration of Mesquite Independent School District for trusting us with this socially impactful project.

\bibliographystyle{splncs04}
\bibliography{mybibliography}

\end{document}